\documentclass[10pt,twocolumn,letterpaper]{article}

\usepackage{iccv}
\usepackage{times}
\usepackage{epsfig}
\usepackage{graphicx}
\usepackage{amsmath}
\usepackage{amssymb}
\usepackage{multirow,makecell,footmisc,url}
\usepackage{subcaption}
\usepackage{gensymb}

\usepackage{comment}

\usepackage[pagebackref=true,breaklinks=true,letterpaper=true,colorlinks,bookmarks=false]{hyperref}

\iccvfinalcopy 

\ificcvfinal\fi
\begin{document}

\title{GCNet: Non-local Networks Meet Squeeze-Excitation Networks and Beyond}

\author{
    Yue Cao$^{1,3*}$, Jiarui Xu$^{2,3}$\thanks{Equal contribution. This work is done when Yue Cao and Jiarui Xu are interns at Microsoft Research Asia.}, Stephen Lin$^3$, Fangyun Wei$^3$, Han Hu$^3$\\
    $^1$School of Software, Tsinghua University\\
    $^2$Hong Kong University of Science and Technology\\
    $^3$Microsoft Research Asia\\
    {\tt \small caoyue10@gmail.com, jxuat@ust.hk, \{stevelin,fawe,hanhu\}@microsoft.com}
}

\maketitle

\begin{abstract}
The Non-Local Network (NLNet) presents a pioneering approach for capturing long-range dependencies, via aggregating query-specific global context to each query position. However, through a rigorous empirical analysis, we have found that the global contexts modeled by non-local network are almost the same for different query positions within an image. In this paper, we take advantage of this finding to create a simplified network based on a query-independent formulation, which maintains the accuracy of NLNet but with significantly less computation. We further observe that this simplified design shares similar structure with Squeeze-Excitation Network (SENet). Hence we unify them into a three-step general framework for global context modeling. Within the general framework, we design a better instantiation, called the global context (GC) block, which is lightweight and can effectively model the global context. The lightweight property allows us to apply it for multiple layers in a backbone network to construct a global context network (GCNet), which generally outperforms both simplified NLNet and SENet on major benchmarks for various recognition tasks.
The code and configurations are released at \url{https://github.com/xvjiarui/GCNet}.
\end{abstract}

\section{Introduction} 
Capturing long-range dependency, which aims to extract the global understanding of a visual scene, is proven to benefit a wide range of recognition tasks, such as image/video classification, object detection and segmentation~\cite{wang2017non,hu2018relation,zhang2018context,hu2018senet}.
In convolution neural networks, as the convolution layer builds pixel relationship in a local neighborhood, the long-range dependencies are mainly modeled by deeply stacking convolution layers.
However, directly repeating convolution layers is computationally inefficient and hard to optimize \cite{wang2017non}.
This would lead to ineffective modeling of long-range dependency, due in part to difficulties in delivering messages between distant positions.

To address this issue, the non-local network \cite{wang2017non} is proposed to model the long-range dependencies using one layer, via self-attention mechanism~\cite{vaswani2017attention}.
For each query position, the non-local network first computes the pairwise relations between the query position and all positions to form an attention map, and then aggregates the features of all positions by weighted sum with the weights defined by the attention map. The aggregated features are finally added to the features of each query position to form the output.

\begin{figure}[]
    \includegraphics[width=1.0\columnwidth]{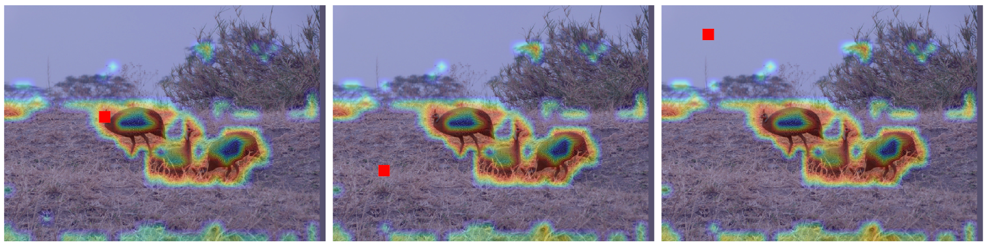}
	\vspace{-20pt}
	\caption{Visualization of attention maps (heatmaps) for different query positions (red points) in a non-local block on COCO object detection. The three attention maps are all almost the same. More examples are in Figure \ref{fig:vis-nlocal}.}
	\label{fig:vis-teaser}
	\vspace{-15pt}
\end{figure}

The query-specific attention weights in the non-local network generally imply the importance of the corresponding positions to the query position. 
While visualizing the query-specific importance weights would help the understanding in depth, such analysis was largely missing in the original paper. We bridge this regret, as in Figure~\ref{fig:vis-teaser}, but surprisingly observe that the attention maps for different query positions are almost the same, indicating only query-independent dependency is learnt. This observation is further verified by statistical analysis in Table~\ref{table:statistical-nlocal} that the distance between the attention maps of different query positions is very small.

Based on this observation, we simplify the non-local block by explicitly using a query-independent attention map for all query positions. Then we add the same aggregated features using this attention map to the features of all query positions for form the output. This simplified block has significantly smaller computation cost than the original non-local block, but is observed with almost no decrease in accuracy on several important visual recognition tasks.
Furthermore, we find this simplified block shares similar structure with the popular Squeeze-Excitation (SE) Network~\cite{hu2018senet}. They both strengthen the original features by the same features aggregated from all positions but differentiate each other by choices on the aggregation strategy, transformation and strengthening functions. By abstracting these functions, we reach a three-step general framework which unifies both the simplified NL block and the SE block: (a) a {context modeling} module which aggregates the features of all positions together to form a global context feature; (b) a feature {transform} module to capture the channel-wise interdependencies; and (c) a {fusion} module to merge the global context feature into features of all positions.

The simplified NL block and SE block are two instantiations of this general framework, but with different implementations of the three steps. By comparison study on each step, we find both the simplified non-local block and the SE block are sub-optimal, that each block has a part of the steps advancing over the other. By a combination of the optimal implementation at each step, we reach a new instantiation of the general framework, called global context (GC) block. The new block shares the same implementation with the simplified NL block on the \emph{context modeling} (using global attention pooling) and \emph{fusion} (using addition) steps, while shares the same \emph{transform} step (using two-layer bottleneck) with SE block. The GC block is shown to perform better than both the simplified non-local block and SE block on multiple visual recognition tasks. 

Like SE block, the proposed GC block is also light-weight which allows it to be applied to all residual blocks in the ResNet architecture, in contrast to the original non-local block which is usually applied after one or a few layers due to  its heavy computation. The GC block strengthened network is named global context network (GCNet). On COCO object detection/segmentation, the GCNet outperforms NLNet and SENet by 1.9\% and 1.7\% on AP${^\text{box}}$, and 1.5\% and 1.5\% on AP${^\text{mask}}$, respectively, with just a 0.07\% relative increase in FLOPs.
In addition, GCNet yields significant performance gains over three general visual recognition tasks: {object detection/segmentation on COCO} (2.7\%$\uparrow$ on AP$^\text{bbox}$, and 2.4\%$\uparrow$ on AP$^\text{mask}$ over Mask R-CNN with FPN and ResNet-50 as backbone \cite{he2017mask}), {image classification on ImageNet} (0.8\%$\uparrow$ on top-1 accuracy over ResNet-50 \cite{he2015resnet}), and {action recognition on Kinetics} (1.1\%$\uparrow$ on top-1 accuracy over the ResNet-50 Slow-only baseline \cite{feichtenhofer2018slowfast}), with less than a 0.26\% increase in computation cost. 

\section{Related Work}
\textbf{Deep architectures.} 
As convolution networks have recently achieved great success in large-scale visual recognition tasks, a number of attempts have been made to improve the original architecture in a bid to achieve better accuracy \cite{krizhevsky2012imagenet,simonyan2014vgg,szegedy2015googlenet,he2015resnet,zagoruyko2016wide,huang2017densely,xie2017resnext,hu2018senet,zoph2018nasnet,hu2018gather,zhang2018shufflenet,howard2017mobilenets,dai2017dcnv1,zhu2018dcnv2,li2018detnet,carreira2017i3d,qiu2017P3D,wang2017non,xie2018rethinking,feichtenhofer2018slowfast}.
An important direction of network design is to improve the functional formulations of basic components to elevate the power of deep networks. 
ResNeXt \cite{xie2017resnext} and Xception \cite{chollet2017xception} adopt group convolution to increase cardinality.
Deformable ConvNets \cite{dai2017dcnv1,zhu2018dcnv2} design deformable convolution to enhance geometric modeling ability.
Squeeze-Excitation Networks \cite{hu2018senet} adopt channel-wise rescaling to explicitly model channel dependencies.

Our global context network is a new backbone architecture, with novel GC blocks to enable more effective global context modeling, offering superior performances on a wide range of vision tasks, such as object detection, instance segmentation, image classification and action recognition.

\textbf{Long-range dependency modeling.} 
The recent approaches for long-range dependency modeling can be categorized into two classes. The first is to adopt self-attention mechanism to model the pairwise relations. The second is to model the query-independent global context.

Self-attention mechanisms have recently been successfully applied in various tasks, such as machine translation \cite{gehring2016convolutional,gehring2017convolutional,vaswani2017attention}, graph embedding \cite{velivckovic2017GAT}, generative modeling \cite{zhang2018SAGAN}, and visual recognition \cite{wang2017residual,hu2018relation,wang2017non,yuan2018ocnet}.
\cite{vaswani2017attention} is one of the first attempts to apply a self-attention mechanism to model long-range dependencies in machine translation. 
\cite{hu2018relation} extends self-attention mechanisms to model the relations between objects in object detection.
NLNet \cite{wang2017non} adopts self-attention mechanisms to model the pixel-level pairwise relations.
CCNet \cite{huang2018ccnet} accelerates NLNet via stacking two criss-cross blocks, and is applied to semantic segmentation.
However, NLNet actually learns query-independent attention maps for each query position, which is a waste of computation cost to model pixel-level pairwise relations.

To model the global context features, SENet \cite{hu2018senet}, GENet \cite{hu2018gather}, and PSANet \cite{zhao2018psanet}  perform rescaling to different channels to recalibrate the channel dependency with global context.
CBAM \cite{woo2018cbam} recalibrates the importance of different spatial positions and channels both via rescaling.
However, all these methods adopt rescaling for feature fusion which is not effective enough for global context modeling.

The proposed GCNet can effectively model the global context via addition fusion as NLNet \cite{wang2017non} (which is heavyweight and hard to be integrated to multiple layers), with the lightweight property as SENet \cite{hu2018senet} (which adopts scaling and is not effective enough for global context modeling).
Hence, via more effective global context modeling, GCNet can achieve better performance than both NLNet and SENet on major benchmarks for various recognition tasks.

\begin{figure*}[!htb]
    \centering
    \includegraphics[width=1.0\linewidth]{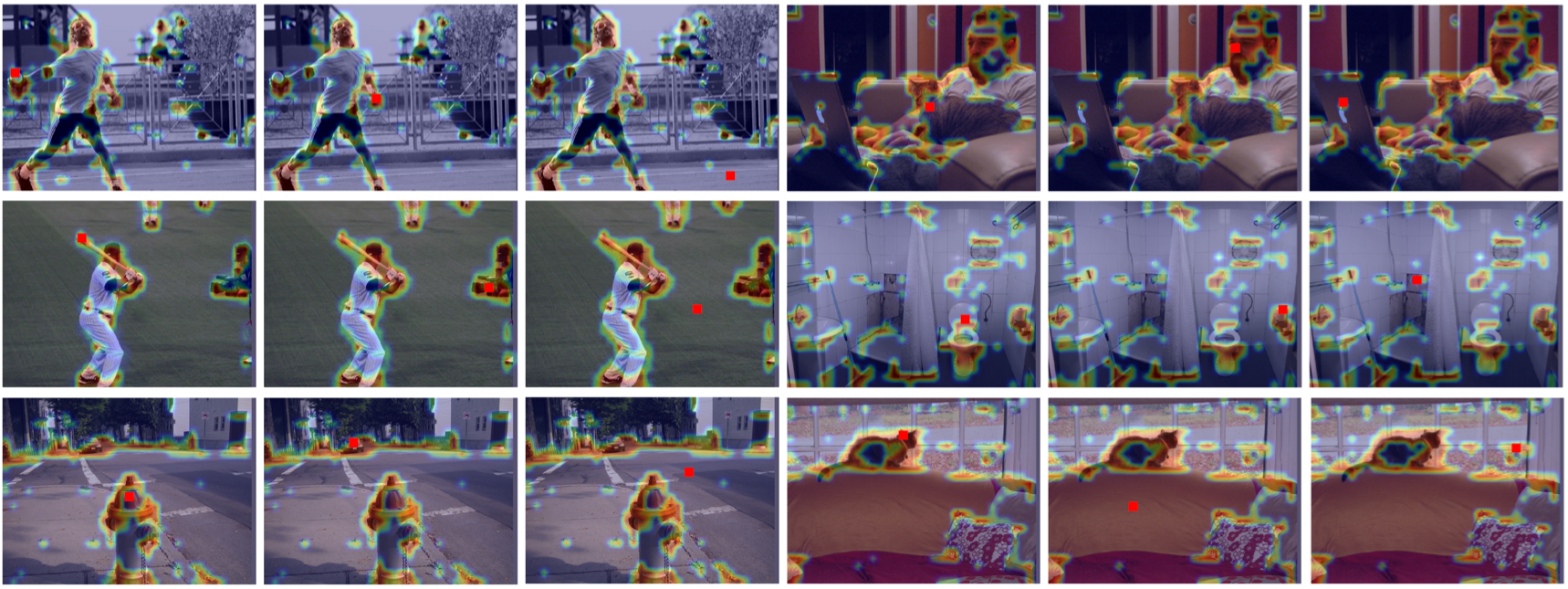}
	\vspace{-8pt}
	\caption{Visualization of attention maps (heatmaps) for different query positions (red points) in a non-local block on COCO object detection. In the same image, the attention maps of different query points are almost the same. \emph{Best viewed in color}.}
	\label{fig:vis-nlocal}
	\vspace{-15pt}
\end{figure*}

\section{Analysis on Non-local Networks}\label{sec:analysis}
In this section, we first review the design of the non-local block \cite{wang2017non}. 
To give an intuitive understanding, we visualize the attention maps across different query positions generated by a widely-used instantiation of the non-local block. 
To statistically analyze its behavior, we average the distances (cosine distance and Jensen-Shannon divergence) between the attention maps of all query positions.

\begin{figure}[!htb]
    \centering
    \includegraphics[width=0.9\columnwidth]{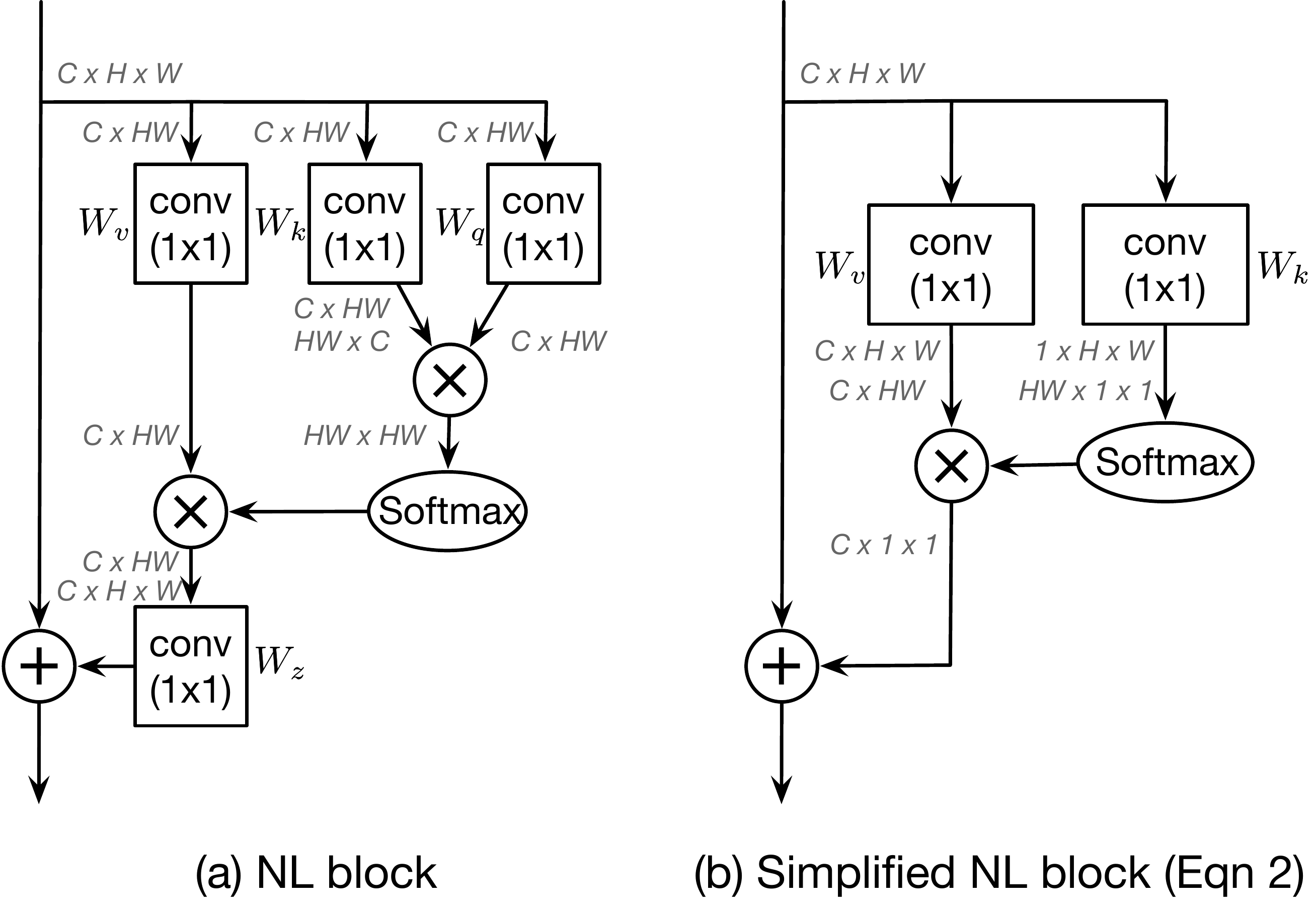}
	\vspace{-8pt}
	\caption{{Architecture of non-local block (Embedded Gaussian) and its simplified version}. The feature maps are shown by their dimensions, e.g. CxHxW. $\otimes$ is matrix multiplication, and $\oplus$ is broadcast element-wise addition. For two matrices with different dimensions, broadcast operations first broadcast features in each dimension to match the dimensions of the two matrices.}
	\label{fig:arch-nl}
	\vspace{-10pt}
\end{figure}

\subsection{Revisiting the Non-local Block} 
The basic non-local block \cite{wang2017non} aims at strengthening the features of the query position via aggregating information from other positions.
We denote ${\bf x}$=$\{{\bf x}_i\}_{i=1}^{N_p}$ as the  feature map of one input instance (e.g., an image or video), where $N_p$ is the number of positions in the feature map (e.g., $N_p$=H$\cdot$W for image, $N_p$=H$\cdot$W$\cdot$T for video). ${\bf x}$ and ${\bf z}$ denote the input and output of the non-local block, respectively, which have the same dimensions. The non-local block can then be expressed as
\begin{equation}
    {{\bf{z}}_i} = {{\bf{x}}_i} + {W_z}\sum\nolimits_{j=1}^{N_p} {\frac{{f\left( {{{\bf{x}}_i},{{\bf{x}}_j}} \right)}}{{{\cal C}\left( {\bf{x}} \right)}}\left( {{W_v} \cdot {{\bf{x}}_j}} \right)},
\end{equation}
where $i$ is the index of query positions, and $j$ enumerates all possible positions. 
$f\left( {{{\bf{x}}_i},{{\bf{x}}_j}} \right)$ denotes the relationship between position $i$ and $j$, and has a normalization factor ${\cal C}\left( {\bf{x}} \right)$.
$W_z$ and $W_v$ denote linear transform matrices (e.g., 1x1 convolution).
For simplification, we denote $\omega_{ij}={\frac{{f\left( {{{\bf{x}}_i},{{\bf{x}}_j}} \right)}}{{{\cal C}\left( {\bf{x}} \right)}}}$ as normalized pairwise relationship between position $i$ and $j$.

To meet various needs in practical applications, four instantiations of the non-local block with different $\omega_{ij}$ are designed, namely Gaussian, Embedded Gaussian, Dot product, and Concat:
(a) Gaussian denotes that $f$ in $\omega_{ij}$ is the Gaussian function, defined as ${\omega _{ij}}$=$\frac{{\exp \left( {\left\langle {{{\bf{x}}_i},{{\bf{x}}_j}} \right\rangle } \right)}}{{\sum\nolimits_m {\exp \left( {\left\langle {{{\bf{x}}_i},{{\bf{x}}_m}} \right\rangle } \right)} }}$;
(b) Embedded Gaussian is a simple extension of Gaussian, which computes similarity in an embedding space, defined as ${\omega _{ij}}$=$\frac{{\exp \left( {\left\langle {{W_q}{{\bf{x}}_i},{W_k}{{\bf{x}}_j}} \right\rangle } \right)}}{{\sum\nolimits_m {\exp \left( {\left\langle {{W_q}{{\bf{x}}_i},{W_k}{{\bf{x}}_m}} \right\rangle } \right)} }}$;
(c) For Dot product, $f$ in $\omega_{ij}$ is defined as a dot-product similarity, formulated as ${\omega _{ij}}$=$\frac{{\left\langle {{W_q}{{\bf{x}}_i},{W_k}{{\bf{x}}_j}} \right\rangle }}{{{N_p}}}$;
(d) Concat is defined literally, as ${\omega _{ij}}$=$\frac{{{\rm{ReLU}}\left( {{W_q}\left[ {{{\bf{x}}_i},{{\bf{x}}_j}} \right]} \right)}}{{{N_p}}}$.
The most widely-used instantiation, Embedded Gaussian, is illustrated in Figure \ref{fig:arch-nl}(a).

The non-local block can be regarded as a global context modeling block, which aggregates query-specific global context features (weighted averaged from all positions via a query-specific attention map) to each query position.
As attention maps are computed for each query position, the time and space complexity of the non-local block are both quadratic to the number of positions $N_p$.

\subsection{Analysis}

\paragraph{Visualization}
To intuitively understand the behavior of the non-local block, we first visualize the attention maps for different query positions.
As different instantiations achieve comparable performance \cite{wang2017non}, here we only visualize the most widely-used version, Embedded Gaussian, which has the same formulation as the block proposed in \cite{vaswani2017attention}.
Since attention maps in videos are hard to visualize and understand, we only show visualizations on the object detection/segmentation task, which takes images as input.
Following the standard setting of non-local networks for object detection \cite{wang2017non}, we conduct experiments on Mask R-CNN with FPN and Res50, and only add one non-local block right before the last residual block of res$_4$.

In Figure \ref{fig:vis-nlocal}, we randomly select six images from the COCO dataset, and visualize three different query positions (red points) and their query-specific attention maps (heatmaps) for each image. 
We surprisingly find that \textbf{for different query positions, their attention maps are almost the same}.
To verify this observation statistically, we analyze the distances between the global contexts of different query positions.

\begin{table}[]
    \centering
    \addtolength{\tabcolsep}{-3.5pt}
    \footnotesize
\begin{tabular}{c|c|cc|ccc|c}
\Xhline{1.0pt}
\multirow{2}{*}{Dataset} & \multirow{2}{*}{Method} & \multirow{2}{*}{AP$^\text{bbox}$} & \multirow{2}{*}{AP$^\text{mask}$}  & \multicolumn{3}{c|}{cosine distance} & \multirow{2}{*}{JSD-att} \\
\cline{5-7}
 &  &  &  & input & output  & att & \\
\hline
\multirow{4}{30pt}{\centering COCO} & Gaussian &  38.0 & 34.8 & 0.397 & 0.062 & 0.177 & 0.065 \\
 & E-Gaussian & 38.0 & 34.7 & 0.402 & 0.012 & 0.020 & 0.011 \\
 & Dot product &  38.1 & 34.8 & 0.405 & 0.020 & 0.015 & - \\
 & Concat &  38.0 & 34.9 & 0.393 & 0.003 & 0.004 & - \\
 \hline
Dataset & Method & Top-1 & Top-5   & input & output  & att & JSD-att \\
 \hline
\multirow{4}{30pt}{\centering Kinetics} & Gaussian & 76.0 & 92.3 & 0.345 & 0.056 & 0.056 & 0.021 \\
 & E-Gaussian & 75.9 & 92.2 & 0.358 & 0.003 & 0.004 & 0.015 \\
 & Dot product & 76.0 & 92.3 & 0.353 & 0.095 & 0.099 & - \\
 & Concat & 75.4 & 92.2 & 0.354 & 0.048 & 0.049 & - \\
\Xhline{1.0pt}
\end{tabular}
	\vspace{-5pt}
    \caption{Statistical analysis on four instantiations of non-local blocks. `input' denotes the input of non-local block (${\bf x}_i$), `output' denotes the output of the non-local block (${\bf z}_i-{\bf x}_i$), `att' denotes the attention map of query positions ($\omega_i$).}
\label{table:statistical-nlocal}
	\vspace{-10pt}
\end{table}

\vspace{-12pt}
\paragraph{Statistical Analysis}
Denote ${\bf v}_i$ as the feature vector for position $i$. The average distance measure is defined as 
$avg\_dist = \frac{1}{{N_p^2}} \sum\nolimits_{i = 1}^{{N_p}} {\sum\nolimits_{j = 1}^{{N_p}} {dist\left( {{{\bf v} _i},{{\bf v} _j}} \right)} }$, 
where $dist(\cdot,\cdot)$ is the distance function between two vectors. 

\textbf{Cosine distance} is a widely-used distance measure, defined as $dist({{\bf v} _i},{{\bf v} _j})$=$(1-\cos({{\bf v} _i},{{\bf v} _j}))/2$.
Here we compute the cosine distance between three kinds of vectors, the non-local block inputs (${\bf v}_i$=${\bf x}_i$, `input' in Table \ref{table:statistical-nlocal}), the non-local block outputs before fusion (${\bf v}_i$=${\bf z}_i$-${\bf x}_i$, `output' in Table \ref{table:statistical-nlocal}), and the attention maps of query positions (${\bf v}_i$=${\omega}_i$, `att' in Table \ref{table:statistical-nlocal}).
The \textbf{Jensen-Shannon divergence} (JSD) is adopted to measure the statistical distance between two probability distributions, as $dist\left( {{{\bf v} _i},{{\bf v} _j}} \right)$=$\frac{1}{2}\sum\nolimits_{k = 1}^{{N_p}} {\left( {v _{ik}}\log \frac{{2{v _{ik}}}}{{{v _{ik}} + {v _{jk}}}} + {v _{jk}}\log \frac{{2{v _{jk}}}}{{{v _{ik}} + {v _{jk}}}}\right)}$. 
As the summation over each attention map $\omega _i$ is 1 (in Gaussian and E-Gaussian), we can regard each ${\omega}_i$ as a discrete probability distribution. Hence we compute JSD between the attention maps (${\bf v}_i$=${\omega}_i$) for Gaussian and E-Gaussian.

Results for two distance measures on two standard tasks are shown in Table \ref{table:statistical-nlocal}.
First, large values of cosine distance in the `input' column show that the input features for the non-local block can be discriminated across different positions. But the values of cosine distance in `output' are quite small, indicating that global context features modeled by the non-local block are almost the same for different query positions.
Both distance measures on attention maps (`att') are also very small for all instantiations, which again verifies the observation from visualization.
In other words, although a non-local block intends to compute the global context specific to each query position, the global context after training is actually independent of query position.
Hence, there is no need to compute query-specific global context for each query position, allowing us to simplify the non-local block.

\begin{figure*}[!htb]
    \centering
    \includegraphics[width=0.95\linewidth]{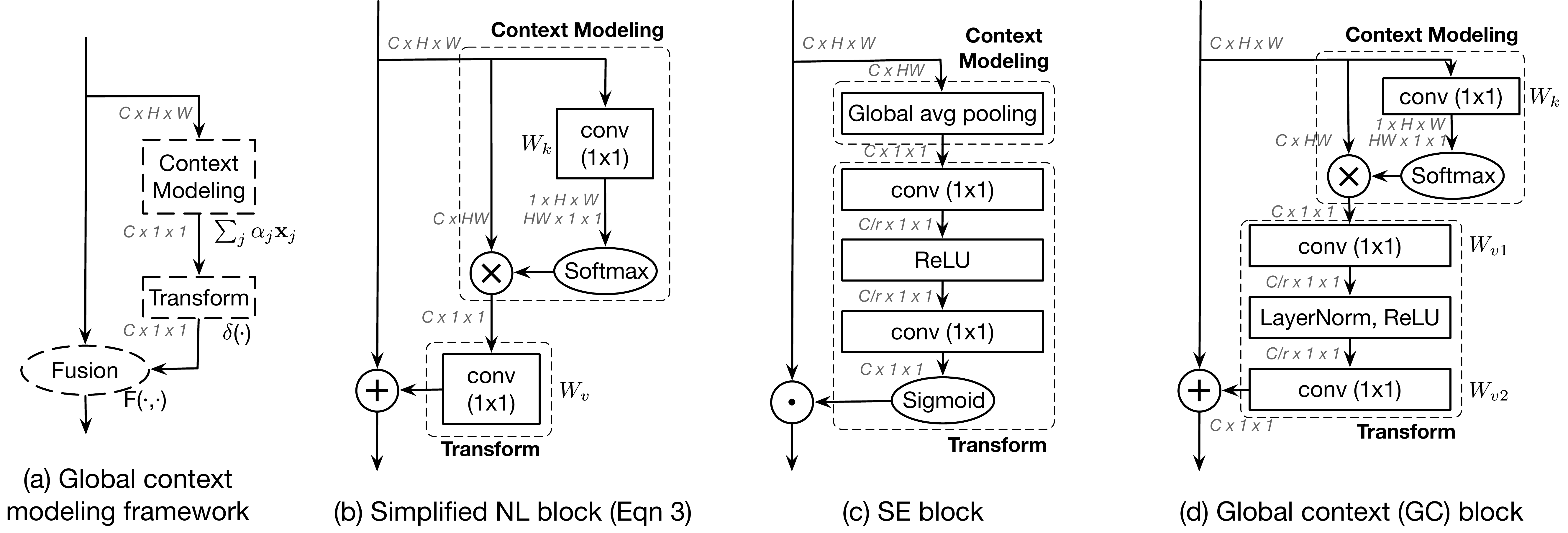}
	\vspace{-5pt}
	\caption{\textbf{Architecture of the main blocks}. The feature maps are shown as feature dimensions, e.g. CxHxW denotes a feature map with channel number C, height H and width W. $\otimes$ denotes matrix multiplication, $\oplus$ denotes broadcast element-wise addition, and $\odot$ denotes broadcast element-wise multiplication.}
	\label{fig:arch}
	\vspace{-10pt}
\end{figure*}

\section{Method}

\subsection{Simplifying the Non-local Block}
As different instantiations achieve comparable performance on both COCO and Kinetics, as shown in Table \ref{table:statistical-nlocal}, here we adopt the most widely-used version, Embedded Gaussian, as the basic non-local block.
Based on the observation that the attention maps for different query positions are almost the same, we simplify the non-local block by computing a global (query-independent) attention map and sharing this global attention map for all query positions. 
Following the results in \cite{hu2018relation} that variants with and without $W_z$ achieve comparable performance, we omit $W_z$ in the simplified version.
Our simplified non-local block is defined as
\begin{equation}\label{eqn:snlocal}
    {{\bf{z}}_i} = {{\bf{x}}_i} + \sum\nolimits_{j=1}^{N_p} {\frac{{\exp \left( {{W_k}{{\bf{x}}_j}} \right)}}{{\sum\nolimits_{m=1}^{N_p} {\exp \left( {{W_k}{{\bf{x}}_m}} \right)} }}\left( {{W_v} \cdot {{\bf{x}}_j}} \right)},
\end{equation}
where $W_k$ and $W_v$ denote linear transformation matrices. 
This simplified non-local block is illustrated in Figure \ref{fig:arch-nl}(b).

To further reduce the computational cost of this simplified block, we apply the distributive law to move $W_v$ outside of the attention pooling, as
\begin{equation}\label{eqn:snlocal2}
    {{\bf{z}}_i} = {{\bf{x}}_i} + {W_v}\sum\nolimits_{j=1}^{N_p} {\frac{{\exp \left( {{W_k}{{\bf{x}}_j}} \right)}}{{\sum\nolimits_{m=1}^{N_p} {\exp \left( {{W_k}{{\bf{x}}_m}} \right)} }}{{\bf{x}}_j}}.
\end{equation}
This version of the simplified non-local block is illustrated in Figure \ref{fig:arch}(b). 
The FLOPs of the 1x1 conv $W_v$ is reduced from $\mathcal{O}$(HWC$^2$) to $\mathcal{O}$(C$^2$).

Different from the traditional non-local block, the second term in Eqn \ref{eqn:snlocal2} is independent to the query position $i$, which means this term is shared across all query positions $i$. 
We thus directly model global context as a weighted average of the features at all positions, and aggregate (add) the global context features to the features at each query position.
In experiments, we directly replace the non-local (NL) block with our simplified non-local (SNL) block, and evaluate accuracy and computation cost on three tasks, object detection on COCO, ImageNet classification, and action recognition, shown in Table \ref{table:ablation-coco}(a), \ref{table:ablation-imagenet}(a) and \ref{table:ablation-kinetics}.
As we expect, the SNL block achieves comparable performance to the NL block with significantly lower FLOPs.

\subsection{Global Context Modeling Framework}
As shown in Figure \ref{fig:arch}(b), the simplified non-local block can be abstracted into three procedures: (a) global attention pooling, which adopts a 1x1 convolution $W_k$ and softmax function to obtain the attention weights, and then performs the attention pooling to obtain the global context features; (b) feature transform via a 1x1 convolution $W_v$; (c) feature aggregation, which employs addition to aggregate the global context features to the features of each position.

We regard this abstraction as a global context modeling framework, illustrated in Figure \ref{fig:arch}(a) and defined as
\begin{equation}\label{eqn:framework}
    {{\bf{z}}_i} = F\left( {{{\bf{x}}_i},\delta \left( {\sum\nolimits_{j=1}^{N_p} {{\alpha _j}{{\bf{x}}_j}} } \right)} \right),
\end{equation}
where (a) ${\sum\nolimits_j {{\alpha _j}{{\bf{x}}_j}} }$ denotes the \textbf{context modeling} module which groups the features of all positions together via weighted averaging with weight $\alpha_j$ to obtain the global context features (global attention pooling in the simplified NL (SNL) block);
(b) $\delta(\cdot)$ denotes the feature \textbf{transform} to capture channel-wise dependencies (1x1 conv in the SNL block); 
and (c) $F(\cdot,\cdot)$ denotes the \textbf{fusion} function to aggregate the global context features to the features of each position (broadcast element-wise addition in the SNL block).

Interestingly, the squeeze-excitation (SE) block proposed in \cite{hu2018senet} is also an instantiation of our proposed framework. 
Illustrated in Figure \ref{fig:arch}(c), it consists of: 
(a) global average pooling for global context modeling (set $\alpha_j=\frac{1}{N_p}$ in Eqn. \ref{eqn:framework}), named the squeeze operation in SE block; 
(b) a bottleneck transform module (let $\delta(\cdot)$ in Eqn. \ref{eqn:framework} be one 1x1 convolution, one ReLU, one 1x1 convolution and a sigmoid function, sequentially), to compute the importance for each channel, named the excitation operation in SE block; and 
(c) a rescaling function for fusion (let $F(\cdot,\cdot)$ in Eqn. \ref{eqn:framework} be element-wise multiplication), to recalibrate the channel-wise features.
Different from the non-local block, this SE block is quite lightweight, allowing it to be applied to all layers with only a slight increase in computation cost.

\subsection{Global Context Block}
Here we propose a new instantiation of the global context modeling framework, named the global context (GC) block, which has the benefits of both the simplified non-local (SNL) block with effective modeling on long-range dependency, and the squeeze-excitation (SE) block with lightweight computation.

In the simplified non-local block, shown in Figure \ref{fig:arch}(b), the transform module has the largest number of parameters, including from one 1x1 convolution with C$\cdot$C parameters.
When we add this SNL block to higher layers, e.g. res$_5$, the number of parameters of this 1x1 convolution, C$\cdot$C=2048$\cdot$2048, dominates the number of parameters of this block.
To obtain the lightweight property of the SE block, this 1x1 convolution is replaced by a bottleneck transform module, which significantly reduces the number of parameters from C$\cdot$C to 2$\cdot$C$\cdot$C/r, where r is the bottleneck ratio and C/r denotes the hidden representation dimension of the bottleneck.
With default reduction ratio set to r=16, the number of params for transform module can be reduced to 1/8 of the original SNL block. More results on different values of bottleneck ratio r are shown in Table \ref{table:ablation-coco}(e).

As the two-layer bottleneck transform increases the difficulty of optimization, we add layer normalization inside the bottleneck transform (before ReLU) to ease optimization, as well as to act as a regularizer that can benefit generalization.
As shown in Table \ref{table:ablation-coco}(d), layer normalization can significantly enhance object detection and instance segmentation on COCO.

The detailed architecture of the global context (GC) block is illustrated in Figure \ref{fig:arch}(d), formulated as
\begin{equation}
\small
{{\bf{z}}_i} = {{\bf{x}}_i} + {W_{v2}}{\rm{ReLU}}\bigg( {{\rm{LN}}\bigg( {{W_{v1}}\sum\limits_{j=1}^{N_p} {\frac{{{e^{{W_k}{{\bf{x}}_j}}}}}{{\sum\limits_{m=1}^{N_p} {{e^{{W_k}{{\bf{x}}_m}}}} }}{{\bf{x}}_j}} } \bigg)} \bigg),
\end{equation}
where $\alpha_j=\frac{{{e^{{W_k}{{\bf{x}}_j}}}}}{{\sum\nolimits_m {{e^{{W_k}{{\bf{x}}_m}}}} }}$ is the weight for global attention pooling, and $\delta(\cdot)=W_{v2}{\rm{ReLU}}({\rm{LN}}(W_{v1}(\cdot)))$ denotes the bottleneck transform.
Specifically, our GC block consists of: (a) global attention pooling for context modeling; (b) bottleneck transform  to capture channel-wise dependencies; and (c) broadcast element-wise addition for feature fusion.

Since the GC block is lightweight, it can be applied in multiple layers to better capture the long-range dependency with only a slight increase in computation cost. Taking ResNet-50 for ImageNet classification as an example, GC-ResNet-50 denotes adding the GC block to all layers (c3+c4+c5) in ResNet-50 with a bottleneck ratio of 16. GC-ResNet-50 increases ResNet-50 computation from $\sim$3.86 GFLOPs to $\sim$3.87 GFLOPs, corresponding to a 0.26\% relative increase.
Also, GC-ResNet-50 introduces $\sim$2.52M additional parameters beyond the $\sim$25.56M parameters required by ResNet-50, corresponding to a $\sim$9.86\% increase.

Global context can benefit a wide range of visual recognition tasks, and the flexibility of the GC block allows it to be plugged into network architectures used in various computer vision problems. In this paper, we apply our GC block to three general vision tasks -- image recognition, object detection/segmentation and action recognition -- and observe significant improvements in all three.

\textbf{Relationship to non-local block.}
As the non-local block actually learns query-independent global context, the global attention pooling of our global context block models the same global context as the NL block but with significantly lower computation cost.
As the GC block adopts the bottleneck transform to reduce redundancy in the global context features, the numbers of parameters and FLOPs are further reduced.
The FLOPs and number of parameters of the GC block are significantly lower than that of NL block, allowing our GC block to be applied to multiple layers with just a slight increase in computation, while better capturing long-range dependency and aiding network training.

\textbf{Relationship to squeeze-excitation block.}
The main difference between the SE block and our GC block is the fusion module, which reflects the different goals of the two blocks. 
The SE block adopts rescaling to recalibrate the importance of channels but inadequately models long-range dependency.
Our GC block follows the NL block by utilizing addition to aggregate global context to all positions for capturing long-range dependency.
The second difference is the layer normalization in the bottleneck transform.
As our GC block adopts addition for fusion, layer normalization can ease optimization of the two-layer architecture for the bottleneck transform, which can lead to better performance. 
Third, global average pooling in the SE block is a special case of global attention pooling in the GC block. 
Results in Table \ref{table:ablation-coco}(f) and \ref{table:ablation-imagenet}(b) show the superiority of our GCNet compared to SENet.

\section{Experiments}
To evaluate the proposed method, we carry out experiments on three basic tasks, object detection/segmentation on COCO \cite{lin2014coco}, image classification on ImageNet \cite{deng2009imagenet}, and action recognition on Kinetics \cite{kay2017kinetics}.
Experimental results demonstrate that the proposed GCNet generally outperforms both non-local networks (with lower FLOPs) and squeeze-excitation networks (with comparable FLOPs).

\subsection{Object Detection/Segmentation on COCO}
We investigate our model on object detection and instance segmentation on COCO 2017 \cite{lin2014coco}, whose train set is comprised of 118k images, validation set of 5k images, and test-dev set of 20k images.
We follow the standard setting \cite{he2017mask} of evaluating object detection and instance segmentation via the standard mean average-precision scores at different
boxes and the mask IoUs, respectively.

\textbf{Setup.}
Our experiments are implemented with PyTorch \cite{paszke2017pytorch}.
Unless otherwise noted, our GC block of ratio $r$=16 is applied to stage c3, c4, c5 of ResNet/ResNeXt.

\textbf{Training.}
We use the standard configuration of Mask R-CNN \cite{he2017mask} with FPN and ResNet/ResNeXt as the backbone architecture.
The input images are resized such that their shorter side is of 800 pixels \cite{he2017fpn}.
We trained on 8 GPUs with 2 images per GPU (effective mini batch size of 16).
The backbones of all models are pretrained on ImageNet classification \cite{deng2009imagenet}, then all layers except for c1 and c2 are jointly finetuned with detection and segmentation heads.
Unlike stage-wise training with respect to RPN in \cite{he2017mask}, end-to-end training like in \cite{ren2015faster} is adopted for our implementation, yielding better results.
Different from the conventional finetuning setting \cite{he2017mask}, we use Synchronized BatchNorm to replace frozen BatchNorm.
All models are trained for 12 epochs using Synchronized SGD with a weight decay of 0.0001 and momentum of 0.9, which roughly corresponds to the 1x schedule in the Mask R-CNN benchmark \cite{massa2018mrcnn}.
The learning rate is initialized to 0.02, and decays by a factor of 10 at the 9th and 11th epochs.
The choice of hyper-parameters also follows the latest release of the Mask R-CNN benchmark~\cite{massa2018mrcnn}.

\begin{table}[]
    \centering
    \footnotesize
    \addtolength{\tabcolsep}{-5.5pt}
\begin{tabular}{c|ccc|ccc|c|c}
\Xhline{1.0pt}
\multicolumn{9}{c}{(a) \textbf{Block design}}                              \\
     & AP${^\text{bbox}}$ & AP$^\text{bbox}_\text{50}$ & AP$^\text{bbox}_{75}$&AP$^\text{mask}$&AP$^\text{mask}_\text{50}$&AP$^\text{mask}_\text{75}$ &  \#param & FLOPs \\
\hline
    baseline & 37.2   & 59.0 & 40.1 & 33.8 & 55.4 & 35.9 & 44.4M & 279.4G \\
+1 NL & 38.0 & 59.8 & 41.0 & 34.7 & 56.7 & 36.6 & 46.5M & 288.7G \\
+1 SNL & 38.1 & 60.0 & 41.6 & 35.0 & 56.9 & 37.0 & 45.4M & 279.4G \\
+1 GC & 38.1  & 60.0 & 41.2 & 34.9 & 56.5 & 37.2 & 44.5M & 279.4G \\
+all GC  & \textbf{39.4} & \textbf{61.6} & \textbf{42.4} & \textbf{35.7} & \textbf{58.4} & \textbf{37.6} & 46.9M & 279.6G \\
\Xhline{1.0pt}
\multicolumn{9}{c}{(b) \textbf{Positions}}                                     \\
     & AP${^\text{bbox}}$ & AP$^\text{bbox}_\text{50}$ & AP$^\text{bbox}_\text{75}$&AP$^\text{mask}$&AP$^\text{mask}_\text{50}$&AP$^\text{mask}_\text{75}$ &  \#param & FLOPs \\
\hline
    baseline & 37.2   & 59.0 & 40.1 & 33.8 & 55.4 & 35.9 & 44.4M & 279.4G \\
afterAdd & \textbf{39.4} & \textbf{61.9} & \textbf{42.5} & \textbf{35.8} & \textbf{58.6} & \textbf{38.1} & 46.9M & 279.6G \\
after1x1 & \textbf{39.4} & 61.6 & 42.4 & 35.7 & 58.4 & 37.6 & 46.9M & 279.6G \\
\Xhline{1.0pt}
\multicolumn{9}{c}{(c) \textbf{Stages}}                                       \\
     & AP${^\text{bbox}}$ & AP$^\text{bbox}_\text{50}$ & AP$^\text{bbox}_\text{75}$&AP$^\text{mask}$&AP$^\text{mask}_\text{50}$&AP$^\text{mask}_\text{75}$ &  \#param & FLOPs \\
\hline
    baseline & 37.2   & 59.0 & 40.1 & 33.8 & 55.4 & 35.9 & 44.4M & 279.4G \\
c3 & 37.9 & 59.6 & 41.1 & 34.5 & 56.3 & 36.8 & 44.5M & 279.5G \\
c4 & 38.9 & 60.9 & 42.2 & 35.5 & 57.6 & 37.7 & 45.2M & 279.5G \\
c5 & 38.7 & 61.1 & 41.7 & 35.2 & 57.4 & 37.4 & 45.9M & 279.4G \\
c3+c4+c5 & \textbf{39.4} & \textbf{61.6} & \textbf{42.4} & \textbf{35.7} & \textbf{58.4} & \textbf{37.6} & 46.9M & 279.6G \\
\Xhline{1.0pt}
\multicolumn{9}{c}{(d) \textbf{Bottleneck design}}                            \\
     & AP${^\text{bbox}}$ & AP$^\text{bbox}_\text{50}$ & AP$^\text{bbox}_\text{75}$&AP$^\text{mask}$&AP$^\text{mask}_\text{50}$&AP$^\text{mask}_\text{75}$ &  \#param & FLOPs \\
\hline
    baseline & 37.2   & 59.0 & 40.1 & 33.8 & 55.4 & 35.9 & 44.4M & 279.4G \\
w/o ratio & \textbf{39.4} & \textbf{61.8} & \textbf{42.8} & \textbf{35.9} & \textbf{58.6} & \textbf{38.1} & 64.4M & 279.6G \\
r16 (ratio 16) & 38.8 & 61.0 & 42.3 & 35.3 & 57.6 & 37.5 & 46.9M & 279.6G \\
r16+ReLU & 38.8 & 61.0 & 42.0 & 35.4 & 57.5 & 37.6 & 46.9M & 279.6G \\
r16+LN+ReLU & \textbf{39.4} & 61.6 & 42.4 & 35.7 & 58.4 & 37.6 & 46.9M & 279.6G \\
\Xhline{1.0pt}
\multicolumn{9}{c}{(e) \textbf{Bottleneck ratio}}                                        \\
     & AP${^\text{bbox}}$ & AP$^\text{bbox}_\text{50}$ & AP$^\text{bbox}_\text{75}$&AP$^\text{mask}$&AP$^\text{mask}_\text{50}$&AP$^\text{mask}_\text{75}$ &  \#param & FLOPs \\
\hline
    baseline & 37.2   & 59.0 & 40.1 & 33.8 & 55.4 & 35.9 & 44.4M & 279.4G \\
ratio 4  & \textbf{39.9} & \textbf{62.2} & \textbf{42.9} & \textbf{36.2} & \textbf{58.7} & \textbf{38.3} & 54.4M & 279.6G \\
ratio 8  & 39.5 & 62.1  & 42.5 & 35.9 & 58.1 & 38.1 & 49.4M & 279.6G \\
ratio 16 & 39.4 & 61.6 & 42.4 & 35.7 & 58.4 & 37.6 & 46.9M & 279.6G \\
ratio 32 & 39.1 & 61.6 & 42.4 & 35.7 & 58.1 & 37.8 & 45.7M & 279.5G \\
\Xhline{1.0pt}
\multicolumn{9}{c}{(f) \textbf{Pooling and fusion}}                                \\
     & AP${^\text{bbox}}$ & AP$^\text{bbox}_\text{50}$ & AP$^\text{bbox}_\text{75}$&AP$^\text{mask}$&AP$^\text{mask}_\text{50}$&AP$^\text{mask}_\text{75}$ &  \#param & FLOPs \\
\hline
    baseline & 37.2   & 59.0 & 40.1 & 33.8 & 55.4 & 35.9 & 44.4M & 279.4G \\
avg+scale (SE) & 38.2 & 60.2 & 41.2 & 34.7 & 56.7 & 37.1 & 46.9M & 279.5G \\
avg+add & 39.1 & 61.4 & 42.3 & 35.6 & 57.9 & \textbf{37.9} & 46.9M & 279.5G \\
att+scale & 38.3 & 60.4 & 41.5 & 34.8 & 57.0 & 36.8 & 46.9M & 279.6G \\
att+add & \textbf{39.4} & \textbf{61.6} & \textbf{42.4} & \textbf{35.7} & \textbf{58.4} & 37.6 & 46.9M & 279.6G \\
\Xhline{1.0pt}
\end{tabular}
	\vspace{-8pt}
\caption{\textbf{Ablation study} based on Mask R-CNN, using ResNet-50 as backbone with FPN, for \textbf{object detection} and \textbf{instance segmentation} on COCO 2017 validation set.}
	\label{table:ablation-coco}
	\vspace{-10pt}
\end{table}

\subsubsection{Ablation Study}
The ablation study is done on COCO 2017 validation set. The standard COCO metrics including AP, AP$_{\text{50}}$, AP$_{\text{75}}$ for both bounding boxes and segmentation masks are reported.

\textbf{Block design.}
Following \cite{wang2017non}, we insert 1 non-local block (NL), 1 simplified non-local block (SNL), or 1 global context block (GC) right before the last residual block of c4.
Table \ref{table:ablation-coco}(a) shows that both SNL and GC achieve performance comparable to NL with fewer parameters and less computation,
indicating redundancy in computation and parameters in the original non-local design.
Furthermore, adding the GC block in all residual blocks yields higher performance (1.1\%$\uparrow$ on AP$^\text{bbox}$ and 0.9\%$\uparrow$ on AP$^\text{mask}$) with a slight increase of FLOPs and \#params.

\textbf{Positions.}
The NL block is inserted after the residual block (afterAdd), while the SE block is integrated after the last 1x1 conv inside the residual block (after1x1).
In Table \ref{table:ablation-coco}(b), we investigate both cases with GC block and they yield similar results. Hence we adopt after1x1 as the default.

\textbf{Stages.}
Table \ref{table:ablation-coco}(c) shows the results of integrating the GC block at different stages.
All stages benefit from global context modeling in the GC block (0.7\%-1.7\%$\uparrow$ on AP$^\text{bbox}$ and AP$^\text{mask}$).
Inserting to c4 and c5 both achieves better performance than to c3, demonstrating that better semantic features can benefit more from the global context modeling.
With slight increase in FLOPs, inserting the GC block to all layers (c3+c4+c5) yields even higher performance than inserting to only a single layer.

\textbf{Bottleneck design.}
The effects of each component in the bottleneck transform are shown in Table \ref{table:ablation-coco}(d).
w/o ratio denotes the simplified NLNet using one 1x1 conv as the transform, which has much more parameters compared to the baseline.
Even though r16 and r16+ReLU have much fewer parameters than the w/o ratio variant, two layers are found to be harder to optimize and lead to worse performance than a single layer.
So LayerNorm (LN) is exploited to ease optimization, leading to performance similar to w/o ratio but with much fewer \#params.

\textbf{Bottleneck ratio.}
The bottleneck design is intended to reduce redundancy in parameters and provide a tradeoff between performance and \#params.
In Table \ref{table:ablation-coco}(e), we alter the ratio r of bottleneck.
As the ratio r decreases (from 32 to 4) with increasing number of parameters and FLOPs, the performance improves consistently (0.8\%$\uparrow$ on AP$^\text{bbox}$ and 0.5\%$\uparrow$ on AP$^\text{mask}$), indicating that our bottleneck strikes a good balance of performance and parameters.
It is worth noting that even with a ratio of r=32, the network still outperforms the baseline by large margins.

\begin{table}[t]
    \footnotesize
    \centering
    \addtolength{\tabcolsep}{-5.5pt}
    \begin{tabular}[t]{cc|ccc|ccc|cc}
    \Xhline{1.0pt}
\multicolumn{9}{c}{(a) \textbf{test on validation set}}                            \\
     \multicolumn{2}{c|}{\centering backbone} &
     AP$^\text{bbox}$ & AP$^\text{bbox}_\text{50}$ & AP$^\text{bbox}_\text{75}$ &
     AP$^\text{mask}$ & AP$^\text{mask}_\text{50}$ & AP$^\text{mask}_\text{75}$ &
     FLOPS \\
    \hline
    \multirow{3}{*}{R50}              & baseline & 37.2 & 59.0 & 40.1 & 33.8 & 55.4 & 35.9 & 279.4G  \\
                                      & +GC r16 & 39.4 & 61.6 & 42.4 & 35.7 & 58.4 & 37.6 &  279.6G \\
                                      & +GC r4& \textbf{39.9} & \textbf{62.2} & \textbf{42.9} & \textbf{36.2} & \textbf{58.7} & \textbf{38.3} & 279.6G \\
    \hline
    \multirow{3}{*}{R101}             & baseline & 39.8 & 61.3 & 42.9 & 36.0 & 57.9 & 38.3 & 354.0G \\
                                      & +GC r16 & 41.1 & 63.6 & 45.0 & 37.4 & 60.1 & 39.6 & 354.3G \\
                                      & +GC r4 & \textbf{41.7} & \textbf{63.7} & \textbf{45.5} & \textbf{37.6} & \textbf{60.5} & \textbf{39.8} & 354.3G \\
    \hline
    \multirow{3}{*}{X101}             & baseline & 41.2 & 63.0 & 45.1 & 37.3 & 59.7 & 39.9 & 357.9G \\
                                      & +GC r16 & 42.4 & 64.6 & 46.5 & 38.0 & 60.9 & 40.5 & 358.2G \\
                                      & +GC r4 & \textbf{42.9} & \textbf{65.2} & \textbf{47.0} & \textbf{38.5} & \textbf{61.8} & \textbf{40.9} & 358.2G \\
    \hline
    \multirow{2}{*}{X101}             & baseline & 44.7 & 63.0 & 48.5 & 38.3 & 59.9 & 41.3 & 536.9G \\
    \multirow{2}{*}{+Cascade}         & +GC r16 & 45.9 & 64.8 & 50.0 & 39.3 & 61.8 & 42.1 & 537.2G \\
                                      & +GC r4 & \textbf{46.5} & \textbf{65.4} & \textbf{50.7} & \textbf{39.7} & \textbf{62.5} & \textbf{42.7} & 537.3G \\

    \hline
    \multirow{2}{*}{X101+DCN}         & baseline & 47.1 & 66.1 & 51.3 & 40.4 & 63.1 & 43.7 & 547.5G \\
    \multirow{2}{*}{+Cascade}         & +GC r16 & \textbf{47.9} & \textbf{66.9} & \textbf{52.2} & \textbf{40.9} & 63.7 & \textbf{44.1} & 547.8G \\
                                      & +GC r4 & \textbf{47.9} & \textbf{66.9} & 51.9 & 40.8 & \textbf{64.0} & 44.0 & 547.8G \\
\hline
\multicolumn{9}{c}{(b) \textbf{test on test-dev set}}  \\
    \hline

    \multirow{2}{*}{X101}             & baseline & 45.0 & 63.7	& 49.1 & 38.7 & 60.8 & 41.8 & 536.9G \\
    \multirow{2}{*}{+Cascade}         & +GC r16 & 46.5 & 65.7 & 50.7 & 40.0 & 62.9 & 43.1 & 537.2G \\
                                      & +GC r4 & \textbf{46.6} & \textbf{65.9} & \textbf{50.7} & \textbf{40.1} & \textbf{62.9} & \textbf{43.3} & 537.3G \\
    \hline
    \multirow{2}{*}{X101+DCN}         & baseline & 47.7 & 66.7 & 52.0 & 41.0 & 63.9 & 44.3 & 547.5G \\
    \multirow{2}{*}{+Cascade}         & +GC r16 & 48.3 & 67.5 & \textbf{52.7} & \textbf{41.5} & \textbf{64.6} & \textbf{45.0} & 547.8G \\
                                      & +GC r4 & \textbf{48.4} & \textbf{67.6} & \textbf{52.7} & \textbf{41.5} & \textbf{64.6} & \textbf{45.0} & 547.8G \\
    \Xhline{1.0pt}
    \end{tabular}
	\vspace{-8pt}
    \caption{Results of GCNet (ratio 4 and 16) with \textbf{stronger backbones} on COCO 2017 validation and test-dev sets.}
	\label{table:archs-coco}
	\vspace{-10pt}
  \end{table}

\textbf{Pooling and fusion.}
The different choices on pooling and fusion are ablated in Table \ref{table:ablation-coco}(f).
First, it shows that addition is more effective than scaling in the fusion stage.
It is surprising that attention pooling only achieves slightly better results than vanilla average pooling.
This indicates that how global context is aggregated to query positions (choice of fusion module) is more important than how features from all positions are grouped together (choice in context modeling module).
It is worth noting that, our GCNet (att+add) significantly outperforms SENet, because of effective modeling of long-range dependency with attention pooling for context modeling, and addition for feature aggregation.

\subsubsection{Experiments on Stronger Backbones}
We evaluate our GCNet on stronger backbones, by replacing ResNet-50 with ResNet-101 and ResNeXt-101 \cite{xie2017resnext}, adding Deformable convolution to multiple layers (c3+c4+c5) \cite{dai2017dcnv1,zhu2018dcnv2} and adopting the Cascade strategy \cite{cai2018cascade}.
The results of our GCNet with GC blocks integrated in all layers (c3+c4+c5) with bottleneck ratios of 4 and 16 are reported.
Table \ref{table:archs-coco}(a) presents detailed results on the validation set.
It is worth noting that even when adopting stronger backbones, the gain of GCNet compared to the baseline is still significant, which demonstrates that our GC block with global context modeling is complementary to the capacity of current models.
For the strongest backbone, with deformable convolution and cascade RCNN in ResNeXt-101, our GC block can still boost performance by 0.8\%$\uparrow$ on AP$^\text{bbox}$ and 0.5\%$\uparrow$ on AP$^\text{mask}$.
To further evaluate our proposed method, the results on the test-dev set are also reported, shown in Table \ref{table:archs-coco}(b). 
On test-dev, strong baselines are also boosted by large margins by adding GC blocks, which is consistent with the results on validation set. 
These results demonstrate the robustness of our proposed method.

\subsection{Image Classification on ImageNet}
ImageNet \cite{deng2009imagenet} is a benchmark dataset for image classification, containing 1.28M training images and 50K validation images from 1000 classes. We follow the standard setting in \cite{he2015resnet} to train deep networks on the training set and report the single-crop top-1 and the top-5 errors on the validation set.
Our preprocessing and augmentation strategy follows the baseline proposed in \cite{xie2018bag} and \cite{hu2018senet}.
To speed up the experiments, all the reported results are trained via two stages.
We first train standard ResNet-50 for 120 epochs on 8 GPUs with 64 images per GPU (effective batch size of 512) with 5 epochs of linear warmup. Second, we insert newly-designed blocks into the model trained in the first stage and finetune for other 40 epochs with a 0.02 initial learning rate. The baseline also follows this two-stage training but without adding new blocks in second stage. Cosine learning rate decay is used for both training and fine-tuning. 

\begin{table}[]
    \footnotesize
    \centering
    \addtolength{\tabcolsep}{-5pt}
\begin{tabular}{c|cc|c|c}
\Xhline{1.0pt}
\multicolumn{5}{c}{(a) \textbf{Block Design}}                                       \\
 & Top-1 Acc & Top-5 Acc & \#params(M) & FLOPs(G) \\
\hline
baseline & 76.88 & 93.16 & 25.56 & 3.86 \\
+1NL & 77.20 & 93.51 & 27.66 & 4.11 \\
+1SNL & 77.28 & 93.60 & 26.61 & 3.86 \\
+1GC & 77.34 & 93.52 & 25.69 & 3.86 \\
+all GC & \textbf{77.70} & \textbf{93.66} & 28.08 & 3.87 \\
\Xhline{1.0pt}
\multicolumn{5}{c}{(b) \textbf{Pooling and fusion}}                                       \\
 & Top-1 Acc & Top-5 Acc & \#params(M) & FLOPs(G) \\
\hline
baseline & 76.88 & 93.16 & 25.56 & 3.86 \\
avg+scale (SENet) & 77.26 & 93.55 & 28.07 & 3.87 \\
avg+add & 77.40 & 93.60 & 28.07 & 3.87 \\
att+scale & 77.34 & 93.48 & 28.08 & 3.87 \\
att+add & \textbf{77.70} & \textbf{93.66} & 28.08 & 3.87 \\
\Xhline{1.0pt}
\end{tabular}
	\vspace{-8pt}
    \caption{\textbf{Ablation study} of GCNet with ResNet-50 on \textbf{image classification} on {ImageNet} validation set.}
	\label{table:ablation-imagenet}
	\vspace{-10pt}
\end{table}

\textbf{Block Design.}
As done for block design on COCO, results on different blocks are reported in Table \ref{table:ablation-imagenet}(a). 
GC block performs slightly better than NL and SNL blocks with fewer parameters and less computation, which indicates the versatility and generalization ability of our design. 
By inserting GC blocks in all residual blocks (c3+c4+c5), the performance is further boosted (by 0.82\%$\uparrow$ on top-1 accuracy compared to baseline) with marginal computational overhead (0.26\% relative increase on FLOPs).

\textbf{Pooling and fusion.}
The functionality of different pooling and fusion methods is also investigated on image classification. Comparing Table \ref{table:ablation-imagenet}(b) with Table \ref{table:ablation-coco}(f), it is seen that attention pooling has greater effect in image classification, which could be one of missing ingredients in \cite{hu2018senet}. Also, attention pooling with addition (GCNet) outperforms vanilla average pooling with scale (SENet) by 0.44\% on top-1 accuracy with almost the same \#params and FLOPs.

\begin{table}[]
    \footnotesize
    \centering
    \addtolength{\tabcolsep}{-2.5pt}
\begin{tabular}{c|cc|c|c}
\Xhline{1.0pt}
method & Top-1 Acc & Top-5 Acc & \#params(M) & FLOPs(G)\\
\hline
baseline & 74.94 & 91.90 & 32.45 & 39.29 \\
\hline
+5 NL    & 75.95 & 92.29 & 39.81 & 59.60 \\
+5 SNL   & 75.76 & \textbf{92.44} & 36.13 & 39.32 \\
+5 GC    & 75.85 & 92.25 & 34.30 & 39.31\\
+all GC  & \textbf{76.00} & 92.34 & 42.45 & 39.35\\
\Xhline{1.0pt}
\end{tabular}
	\vspace{-8pt}
    \caption{Results of GCNet and NLNet based on Slow-only baseline using R50 as backbone on \textbf{Kinetics} validation set.}
	\label{table:ablation-kinetics}
	\vspace{-10pt}
\end{table}

\subsection{Action Recognition on Kinetics}
For human action recognition, we adopt the widely-used Kinetics \cite{kay2017kinetics} dataset, which has $\sim$240k training videos and 20k validation videos in 400 human action categories. 
All models are trained on the training set and tested on the validation set. 
Following \cite{wang2017non}, we report top-1 and top-5 recognition accuracy.
We adopt the slow-only baseline in \cite{feichtenhofer2018slowfast}, the best single model to date that can utilize weights inflated \cite{carreira2017i3d} from the ImageNet pretrained model.
This inflated 3D strategy \cite{wang2017non} greatly speeds up convergence compared to training from scratch. 
All the experiment settings explicitly follow \cite{feichtenhofer2018slowfast}; the slow-only baseline is trained with 8 frames ($8\times8$) as input, and multi(30)-clip validation is adopted.

The ablation study results are reported in Table \ref{table:ablation-kinetics}. 
For Kinetics experiments, the ratio of GC blocks is set to 4. 
First, when replacing the NL block with the simplified NL block and GC block, the performance can be regarded as on par (0.19\%$\downarrow$ and 0.11\%$\downarrow$ in top-1 accuracy, 0.15\%$\uparrow$ and 0.14\%$\uparrow$ in top-5 accuracy). 
As in COCO and ImageNet, adding more GC blocks further improves results and outperforms NL blocks with much less computation.

\section{Conclusion}
The pioneering work for long-range dependency modeling, non-local networks, intends to model query-specific global context, but only models query-independent context. Based on this, we simplify non-local networks and abstract this simplified version to a global context modeling framework. Then we propose a novel instantiation of this framework, the GC block, which is lightweight and can effectively model long-range dependency.
Our GCNet is constructed via applying GC blocks to multiple layers, which generally outperforms simplified NLNet and SENet on major benchmarks for various recognition tasks.

{
\bibliographystyle{ieee}
\bibliography{SalNet}
}

\end{document}